# Overview of CHIP 2026 Shared Task 1: Patient-Specific Phenotype Concept Recognition in Full-Text Biomedical Literature

Xuanlyu Zhu[1], Feng Tian[1], Yang Zhong[1], Pengqi Zheng[1], Xiwen Bi[2], Ling Luo[1] (✉), Yuanyuan Sun[1], Jian Wang[1] and Hongfei Lin[1]

[1]Dalian University of Technology, Dalian 116023, Liaoning, China

[2]Xiangya School of Medicine, Central South University, Changsha 410013, Hunan, China

lingluo@dlut.edu.cn

## Abstract

Automatic phenotype information extraction plays an important role in rare disease diagnosis, genotype-phenotype analysis, and precision medicine. However, accurately identifying phenotype concepts and determining their associations with specific patients remain challenging, particularly in full-text biomedical articles involving multiple patients and complex clinical descriptions. This paper presents an overview of CHIP 2026 Shared Task 1, which aims to promote the development of methods for patient-specific phenotype concept recognition from full-text biomedical literature. The task consists of two subtasks: full-text phenotype concept recognition and patient-specific phenotype association. For this evaluation, we used 200 annotated full-text biomedical articles from the PatientPheX dataset, containing 506 patients, 15,193 phenotype mentions, and 5,656 patient-phenotype associations. A total of 455 teams registered for the competition, with 189 and 103 valid submissions received on the Phase A and Phase B leaderboards, respectively. The best-performing system achieved an overall score of 0.7389 on the final test set, outperforming the baseline by 8.37 percentage points. The top-performing teams explored diverse approaches combining biomedical language models, large language models (LLMs), and ensemble strategies. The evaluation results highlight both the potential of these approaches and the remaining challenges in patient attribution and phenotype normalization from full-text biomedical literature. The post-evaluation phase remains open at https://tianchi.aliyun.com/competition/entrance/532512.

## 1 Introduction

Human phenotypes refer to observable or measurable clinical characteristics resulting from the interaction between genetic and environmental factors, such as height, blood pressure, combinations of symptoms, and drug responses. Phenotypic information plays an important role in rare disease diagnosis, disease mechanism investigation, gene discovery, and precision medicine [1-3]. The Human Phenotype Ontology (HPO) provides a standardized vocabulary for describing phenotypic abnormalities associated with human diseases, enabling the systematic representation and analysis of clinical phenotype [1]. By identifying phenotype mentions in biomedical text and normalizing them to HPO concepts, unstructured clinical descriptions can be transformed into standardized phenotype information, supporting the construction of patient-specific phenotype profiles and downstream applications such as phenotype-driven gene prioritization and genetic diagnosis [2, 4]. With the rapid growth of biomedical literature, automatically extracting, normalizing, and organizing phenotype information from large volumes of unstructured text has become an important research direction in biomedical natural language processing.

Phenotype Concept Recognition (PCR) aims to automatically identify phenotype mentions in biomedical text and normalize them to standardized concepts in HPO. In recent years, various approaches have been developed for this task, including dictionary-based methods, machine learning models, pretrained language models, and large language models [5-8]. Despite these advances, accurate phenotype recognition and normalization remain challenging. Biomedical texts contain diverse phenotype expressions, including abbreviations, descriptive mentions, negated findings, and composite phenotypes, which increase the difficulty of identifying entity boundaries and determining their corresponding HPO identifiers [5, 7]. Moreover, the large number of concepts and complex hierarchical structure of HPO make accurate concept normalization difficult, particularly when phenotype descriptions differ from standardized ontology terms [1, 6]. These challenges become more pronounced in full-text biomedical articles, where phenotype information may be distributed across multiple sections and described in different contexts.

Existing studies and evaluation resources have primarily focused on identifying and normalizing phenotype concepts from biomedical abstracts, clinical observations, or other relatively short texts [9-11]. Although these resources have contributed to the development of phenotype recognition methods, they provide limited support for patient-specific phenotype information extraction from full-text biomedical literature. In particular, biomedical case reports and case series may describe multiple patients, family members, or other individuals within the same article, each presenting different clinical phenotype. Therefore, identifying all phenotype concepts mentioned in a document does not necessarily indicate which phenotype are associated with a specific patient. Models must further distinguish patient-specific findings from those associated with other individuals, as well as negated or background information. Such patient attribution is essential for constructing accurate patient-specific phenotype profiles and supporting genotype-phenotype association analysis and personalized medical research [2, 4].

To address these challenges, we organized the shared task of Patient-specific Phenotype Concept Recognition from Full-text Biomedical Literature at the China Health Information Processing Conference (CHIP 2026). The task aims to promote the development and evaluation of methods for extracting phenotype information associated with specific patients from full-text biomedical articles. As illustrated in Fig. 1, the task consists of two subtasks: (1) Full-text Phenotype Concept Recognition, which requires systems to identify all phenotype mentions in a biomedical article and normalize them to HPO concepts; and (2) Patient-specific Phenotype Association, which requires systems to identify the set of phenotype concepts associated with a given patient. The evaluation was conducted using the PatientPheX dataset [12], comprising 200 full-text biomedical articles from PubMed Central (PMC) [13], with annotations of phenotype mentions, HPO identifiers, and patient-phenotype associations. By combining phenotype recognition and patient attribution, this shared task provides a standardized benchmark for evaluating patient-specific phenotype information extraction from full-text biomedical literature.

The evaluation competition was hosted on the Tianchi platform (https://tianchi.aliyun.com/competition/entrance/532512) and consisted of two phases: Phase A (development set) for preliminary evaluation and Phase B (test set) for final leaderboard ranking. The competition attracted active participation, with a total of 455 teams registered, 189 valid submissions to the Phase A leaderboard, and 103 valid submissions to the Phase B leaderboard. The best-performing systems achieved overall scores of 0.8071 and 0.7389 in Phase A and Phase B, respectively, outperforming the corresponding baselines. The evaluation results also revealed that patient-specific phenotype association remains more challenging than full-text phenotype concept recognition, highlighting the need for further research on patient attribution and document-level information extraction. To further advance research in the field, a post-evaluation phase was launched after the competition to encourage continued exploration using the dataset.

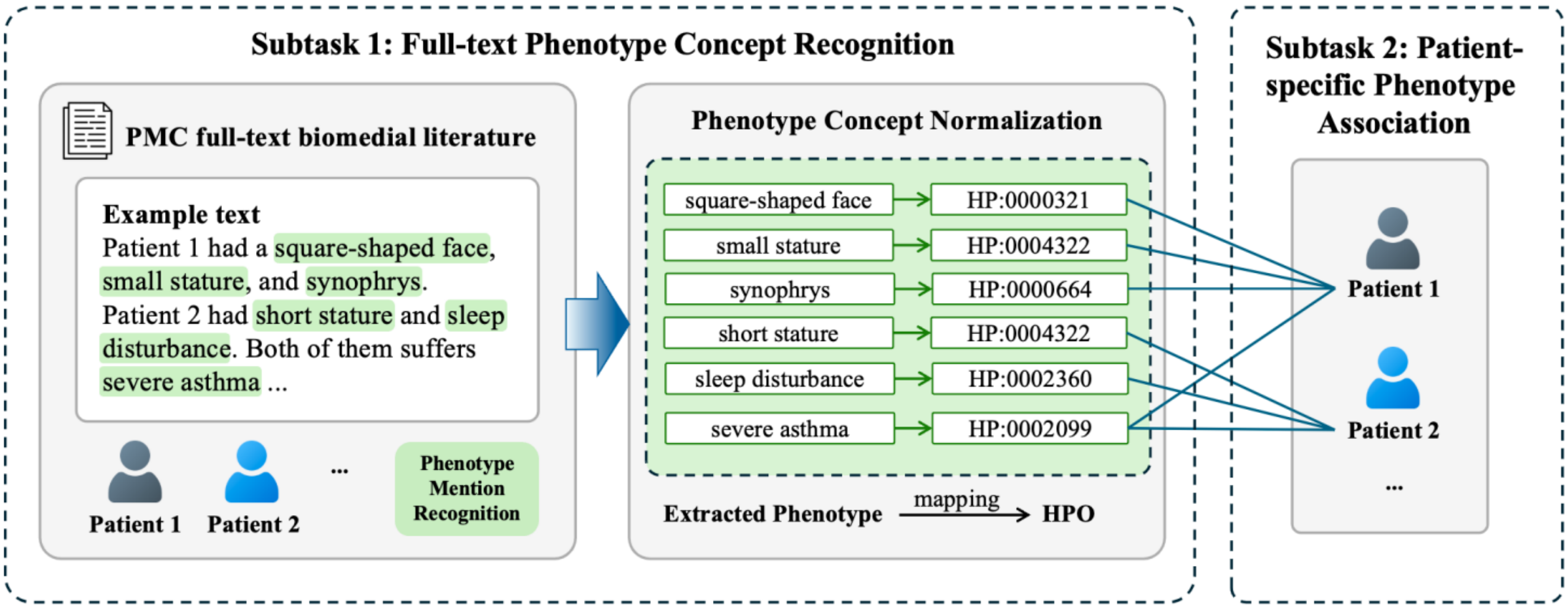

Fig. 1. Overview of the two subtasks in the CHIP 2026 shared task

## 2 Related Work

To promote the development of biomedical natural language processing, various shared tasks and evaluation challenges have been organized, including BioCreative [14], the China Conference on Knowledge Graph and Semantic Computing (CCKS) [15], and CHIP. These evaluations cover a wide range of biomedical information extraction tasks, such as named entity recognition, concept normalization, and relation extraction. Among them, BioCreative VIII Task 3 specifically focused on genetic phenotype normalization from dysmorphology physical examinations [10]. The task provided 3,136 manually annotated clinical observations from 1,652 pediatric patients and required participating systems to identify phenotype mentions and normalize them to HPO concepts. However, this evaluation primarily focused on relatively short clinical observations, while phenotype information extraction from full-text biomedical articles and the attribution of phenotype concepts to specific patients remain challenging.

In recent years, various methods have been developed for phenotype concept recognition and patient-specific information extraction. For phenotype recognition and normalization, PhenoTagger [5] proposed a hybrid framework combining dictionary-based matching and deep learning models, while PhenoTagger+ [16] further incorporated rich HPO information through LLM-based synonym augmentation and ontology-enhanced concept representations to improve phenotype recognition accuracy and robustness. PhenoBERT [6] integrated convolutional neural networks (CNNs) for candidate HPO concept selection with BERT-based semantic matching. FastHPOCR [7] introduced an efficient dictionary-based approach to address lexical variations in phenotype expressions, and AutoPCR [8] further explored a prompt-based framework combining phenotype mention extraction, semantic concept retrieval, and LLM-based entity linking. PhenoBCBERT and PhenoGPT [17] investigated pretrained and generative language models for phenotype recognition. For patient-specific phenotype extraction, document-level relation extraction methods provide potential solutions by capturing relationships between entities across sentences. ATLOP [18] introduced adaptive thresholding and localized context pooling to improve relation extraction performance, while GLiM [19] combined a graph-enhanced Transformer with LLMs to address incomplete annotations in biomedical relation extraction. Despite these advances, accurately identifying phenotype concepts and determining their associations with specific patients remain challenging, particularly in full-text biomedical articles involving multiple patients or family members.

Several annotated corpora have also been developed to support the training and evaluation of phenotype extraction methods. The GSC corpus and its updated version, GSC 2024, provide phenotype mention annotations and corresponding HPO concepts in biomedical abstracts [9, 7], while ID-68 focuses on clinical case descriptions [6]. Beyond short texts, the JAX corpus provides document-level HPO annotations for full-text biomedical articles [5], and the Case Study Cohort (CSC) contains manually annotated phenotype concepts derived from published clinical cases [11]. However, existing resources provide limited support for jointly evaluating full-text phenotype recognition, HPO normalization, and patient-specific phenotype associations. Therefore, we introduce PatientPheX, a patient-centered full-text biomedical dataset annotated with phenotype mentions, standardized HPO concepts, and explicit patient–phenotype associations [12]. For the CHIP 2026 shared task, 200 annotated full-text articles from PatientPheX were used to evaluate full-text phenotype concept recognition and patient-specific phenotype information extraction.

## 3 Material and Methods

### 3.1 Task Definition

The shared task aims to evaluate and compare the performance of participating teams in patient-specific phenotype information extraction from full-text biomedical articles publicly available in PMC. Given a full-text article, participating teams are required to identify phenotype concepts and determine their associations with specific patients. The task consists of two subtasks:

- **Subtask1:** Full-text Phenotype Concept Recognition. Participating teams are required to identify all phenotype mentions in a full-text biomedical article, determine their corresponding text spans, and normalize them to standardized concepts in the HPO. All phenotype concepts are restricted to the Phenotypic abnormality (HP:0000118) branch of HPO version 2026-06-23. This subtask focuses on phenotype recognition and normalization without considering patient attribution.

- **Subtask2:** Patient-specific Phenotype Association. Given a full-text biomedical article and a specified patient, participating teams are required to identify the phenotype concepts associated with that patient and return a set of corresponding HPO identifiers. Unlike Subtask 1, this subtask focuses on patient attribution, requiring systems to distinguish phenotype associated with the target patient from those belonging to other individuals mentioned in the same article.

Participants were encouraged to explore diverse technical approaches, including traditional machine learning, deep learning, pretrained language models, and large language models. Methods such as prompt learning, supervised fine-tuning, and retrieval-augmented generation (RAG) were also permitted. To ensure fairness and comparability, the task imposed several constraints: (1) Each individual model was limited to 10 billion parameters; (2) The use of external manually annotated phenotype datasets was prohibited; and (3) All models and external knowledge resources used by participating teams were required to be explicitly described in their technical reports. These requirements aimed to encourage methodological diversity while ensuring a fair and standardized evaluation of participating systems.

### 3.2 Dataset Description

The evaluation was conducted using the PatientPheX dataset [12], a full-text biomedical corpus developed for phenotype concept recognition and patient-specific phenotype information extraction. The dataset consists of 200 full-text biomedical articles collected from PMC, with annotations of phenotype mentions, standardized HPO concepts, and patient-phenotype associations. In total, PatientPheX contains 506 patients, 15,193 phenotype mentions, and 2,874 unique HPO concepts. It also includes complex phenotype expressions, such as negated and composite phenotype, to support the evaluation of systems in realistic biomedical text.

The dataset is organized in JSON Lines (JSONL) format, with each line representing a full-text article. Each document contains five main fields: pmc_id and pmid provide the article identifiers; patient records the target patients and their textual mentions; full_text contains the article content organized into sections; entities provides phenotype mentions with their text spans and corresponding HPO identifiers; and association specifies the set of phenotype concepts associated with each patient. Text spans are represented using character-level offsets and lengths relative to the original full text. For phenotype normalization, composite phenotype mentions may be mapped to multiple HPO identifiers, separated by semicolons, while mentions without valid HPO mappings are assigned -1. Negated phenotype mentions are additionally marked with NO. An example of the dataset format is presented in Table 1.

Table 1. Example data format

| Field | Subfield | Example |
|---|---|---|
| pmc_id | - | 12652105 |
| pmid | - | 41300704 |
| patient | patient_id | OII.1 |
| | mention.text | patient one (II.1) |
| | mention.offset | 12645 |
| | mention.length | 18 |
| full_text | section_type | TITLE |
| | type | front |
| | offset | 0 |
| | text | Rare Duplication in the RYR1 Gene... |
| entities | identifier | HP:0002047 |
| | type | Phenotype |
| | text | Malignant Hyperthermia |
| | offset | 42 |
| | length | 22 |
| | note | null |
| association | patient_id | OII.1 |
| | phenotype | ["HP:0003938", "HP:0001945", ...] |

For model development and evaluation, the dataset was divided into training, development, and test sets, containing 80, 20, and 100 articles, respectively. The training set provides annotated examples for model training, while the development and test sets correspond to the A and B leaderboards of the shared task. Gold-standard annotations for the development and test sets were withheld from participants, who were required to submit their predictions through the online evaluation platform. In addition to 15,193 phenotype mentions, the dataset contains 435 negated phenotype mentions, 212 composite phenotype mentions, and 5,656 patient-phenotype associations. Detailed statistics of each subset are presented in Table 2.

Table 2. Statistic summary of the PatientPheX dataset

| Statistics | Train | Dev (A) | Test (B) | Total |
|---|---|---|---|---|
| Documents | 80 | 20 | 100 | 200 |
| Patients | 209 | 53 | 244 | 506 |
| Phenotype mentions | 6,027 | 1,563 | 7,603 | 15,193 |
| Unique HPO concepts | 1,662 | 551 | 1,863 | 2,874 |
| Negated phenotypes | 171 | 38 | 226 | 435 |
| Composite phenotypes | 69 | 5 | 138 | 212 |
| Patient-phenotype associations | 2,307 | 563 | 2,786 | 5,656 |
| Unique patient-associated HPO concepts | 1,091 | 351 | 1,252 | 1,995 |

**3.3 Baseline System**

To provide a reference for evaluating participating systems, we developed a pipeline-based baseline system consisting of phenotype concept recognition and patient-specific phenotype association, corresponding to the two subtasks.

For Subtask 1, PhenoTagger [5] was employed to identify phenotype mentions from full-text biomedical articles and normalize them to standardized HPO concepts using the designated HPO version (2026-06-23). The recognition results, including phenotype text spans and corresponding HPO identifiers, were used for both mention-level and document-level evaluation and served as candidate phenotype entities for Subtask 2.

Stage 1: Patient mention recognition
You are given a passage from a biomedical article and a list of target patients. Each target patient is represented by a patient identifier and one or more original mentions.
Your task is to identify all textual mentions in the passage that refer to one of the specified patients.
Target patients:
{patient_information}
Passage offset in the full article:
{passage_offset}
Passage:
{passage_text}
Requirements:
1. Only return mentions referring to the supplied patient identifiers.
2. A mention may be an explicit patient identifier, case number, pedigree label, patient description, or an unambiguous coreferential expression.
3. Do not assign relatives, controls, authors, generic patient groups, or other individuals to a target patient.
4. Assign pronouns or generic descriptions only when the referent is clear from the local context.
5. The text span must exactly match the source text.
6. Offset values must be calculated relative to the full article.
7. If a mention is ambiguous, omit it.
8. Return at most 25 mentions for each patient.
9. Return JSON only.
Output format:
{"patients": [
{"patient_id": "O1",
"mentions": [{
"text": "the proband",
"offset": 1234,
"length": 11}]
}]}
Stage 2: Patient-phenotype association
You are given a biomedical passage, a list of target patients and their resolved mentions, and phenotype entities recognized by PhenoTagger.
Your task is to determine which phenotype entities are associated with each target patient.
Target patients and resolved mentions:
{resolved_patient_mentions}
Candidate phenotype entities:
{phenotype_candidates}
Passage or contextual evidence:
{context}
Requirements:
1. Assign only the supplied phenotype entity indices to the supplied patient identifiers.
2. Use patient mentions, sentence-level evidence, and surrounding context to determine phenotype ownership.
3. A phenotype may be assigned to multiple patients only when the text explicitly supports all corresponding associations.
4. Exclude phenotypes referring only to relatives, controls, cohorts, literature cases, animals, or cell lines.
5. Exclude negated, hypothetical, ruled-out, and family-history-only phenotypes.
6. Do not infer a phenotype solely from general disease knowledge.
7. Omit associations when the patient identity or phenotype ownership is ambiguous.
8. Return an empty entity list when no candidate is supported.
9. Return JSON only.
Output format:
{"assignments": [
{"patient_id": "O1",
"entity_indices": [0, 3, 7]},
{"patient_id": "O2",
"entity_indices": []}
]}

Fig. 2. Prompt templates for the two-stage prompting strategy in Subtask 2

For Subtask 2, we employed Qwen3-8B [20] in a zero-shot setting to identify patient–phenotype associations through a two-stage prompting strategy. In the first stage, each article was divided into passages, and the model was provided with predefined patient identifiers, their original mentions, and the corresponding textual context to identify additional mentions referring to each target patient. These mentions included explicit patient identifiers, case labels, pedigree identifiers, and unambiguous coreferential expressions. In the second stage, the identified patient mentions and phenotype entities recognized by PhenoTagger were provided to the model, together with their surrounding contexts, to determine the associations between patients and phenotype concepts. The contextual information for each phenotype candidate included up to 500 characters on either side, and phenotype candidates were processed in batches of eight. The prompt templates for both stages are shown in Fig. 2.

During post-processing, predictions involving invalid patient identifiers or phenotype candidates were removed, and duplicate associations were eliminated. Negated, hypothetical, and ambiguous phenotype associations were also excluded. The final output consisted of a set of HPO identifiers associated with each target patient.

The baseline system did not use the task training set for supervised fine-tuning or parameter optimization. For Subtask 2, Qwen3-8B was used with greedy decoding (temperature = 0), with maximum output lengths of 1,800 and 2,600 tokens for patient mention recognition and phenotype association, respectively. All predictions were generated through zero-shot prompting without parameter updates. This baseline provides a reproducible reference for evaluating the effectiveness of different phenotype recognition and patient-specific association methods developed by participating teams.

### 3.4 Evaluation Matric

To comprehensively evaluate the performance of participating systems, precision (P), recall (R), and F1 score were adopted for both subtasks. For Subtask 1, mention-level and document-level evaluations were used to assess phenotype recognition and normalization. For Subtask 2, micro-averaged and macro-averaged F1 scores were employed to evaluate patient-specific phenotype associations. These metrics share the same basic definitions of precision, recall, and F1, but differ in how true positives (TP), false positives (FP), and false negatives (FN) are determined and aggregated:

$$P = \frac{TP}{TP+FP}, \quad R = \frac{TP}{TP+FN}, \quad F1 = \frac{2PR}{P+R} \qquad (1)$$

**Subtask 1: Full-text Phenotype Concept Recognition.** This subtask evaluates system performance at both the mention and document levels.

At the mention level, a predicted phenotype is considered correct only when its text span exactly matches the gold-standard annotation and its corresponding HPO identifier is correctly assigned. Each correctly matched phenotype mention is counted as a TP, while unmatched predictions and gold-standard annotations are counted as FP and FN, respectively. Mention-level precision, recall, and F1 are calculated using Eq. (1), based on the aggregated counts across all test documents.

At the document level, the evaluation focuses on identifying the set of phenotype concepts mentioned in each article, regardless of their text spans or occurrence frequencies. Let $G_d$ and $P_d$ denote the sets of unique gold-standard and predicted HPO identifiers in document \(d\), respectively. Document-level TP, FP, and FN are aggregated across all $N$ test documents as follows:

$$\begin{aligned} TP_{\text{doc}} &= \textstyle\sum_{d=1}^{N} |\, G_d \cap P_d \,|, \\ FP_{\text{doc}} &= \textstyle\sum_{d=1}^{N} |\, P_d \setminus G_d \,|, \\ FN_{\text{doc}} &= \textstyle\sum_{d=1}^{N} |\, G_d \setminus P_d \,|. \end{aligned} \quad (2)$$

Document-level precision, recall, and F1 are then calculated using Eq. (1).

Special annotation cases were handled according to predefined evaluation rules. Negated phenotype mentions marked with NO were excluded from the gold-standard positive instances. Correctly identified negated mentions were also excluded from scoring, whereas negated mentions predicted as positive phenotypes were counted as false positives. For composite phenotype mentions mapped to multiple HPO identifiers, each identifier was treated as an independent evaluation unit. For phenotype mentions without valid HPO mappings (assigned -1), correct text span identification was sufficient for a match.

**Subtask 2: Patient-specific Phenotype Association.** This subtask evaluates the correctness of patient–phenotype associations using micro-averaged and macro-averaged F1 scores. A predicted patient–phenotype association is considered correct only when the phenotype concept is correctly assigned to the specified patient. Incorrectly predicted associations are counted as FP, while missing gold-standard associations are counted as FN.

Micro-F1 measures the overall performance by aggregating TP, FP, and FN across all patients in the test set and applying Eq. (1).

Macro-F1 evaluates performance at the individual patient level by assigning equal weight to each patient. Let $y_i$ and $\hat{y}_i$ denote the sets of gold-standard and predicted phenotype identifiers for patient *i*, respectively. The TP, FP, and FN for each patient are determined as follows:

$$\begin{aligned} TP_i &= |y_i \cap \hat{y}_i|, \\ FP_i &= |\hat{y}_i \setminus y_i|, \\ FN_i &= |y_i \setminus \hat{y}_i|. \end{aligned} \quad (3)$$

The precision, recall, and F1 for each patient are calculated using Eq. (1). The macro-averaged F1 is then obtained by averaging the individual F1 scores across all *K* gold-standard patients:

$$\mathrm{F1}_{\text{macro}} = \frac{1}{K}\textstyle\sum_{i=1}^{K} \mathrm{F1}_i \quad (4)$$

When both the gold-standard and predicted phenotype sets for a patient are empty, the corresponding precision, recall, and F1 are defined as 1.

**Overall Evaluation Score.** The final leaderboard score combines the performance of both subtasks using a weighted sum of the four F1 metrics:

$$\text{Score} = 0.25(F1_{\text{men}} + F1_{\text{doc}}) + 0.25(F1_{\text{micro}} + F1_{\text{macro}}) \quad (12)$$

This composite score assigns equal weight to the two subtasks, balancing phenotype recognition and normalization performance with patient-specific phenotype association accuracy.

## 4 Results

A total of 455 teams registered for this shared task. During the competition, 189 valid submissions were received on the Phase A leaderboard, and 103 valid submissions were received on the Phase B leaderboard. Participating teams were allowed to submit up to five runs per day during Phase A and three runs per day during Phase B. These submission limits were established to ensure fair competition and encourage participants to develop and optimize their methods.

### 4.1 Evaluation Results

During Phase A, participants were evaluated on the development set, while Phase B used the test set for the final leaderboard ranking. The best results achieved by participating teams, along with the baseline performances in both phases, are summarized in Table 3.

Table 3. Best results and baseline performances on both leaderboards

| Phase | Type | Men-F1 | Doc-F1 | Micro-F1 | Macro-F1 | Score |
|---|---|---|---|---|---|---|
| A | Best result | 0.7891 | 0.8531 | 0.7783 | 0.8077 | 0.8071 |
| | Baseline | 0.7382 | 0.7947 | 0.6797 | 0.6878 | 0.7251 |
| B | Best result | 0.7719 | 0.8286 | 0.7055 | 0.6496 | 0.7389 |
| | Baseline | 0.7211 | 0.7933 | 0.5874 | 0.5188 | 0.6552 |

As shown in Table 3, the best-performing systems achieved notable improvements over the baseline across both phases. The highest overall scores in Phase A and Phase B were 0.8071 and 0.7389, outperforming the corresponding baselines by 8.20 and 8.37 percentage points, respectively. In Phase B, the best-performing system achieved a mention-level F1 of 0.7719 and a document-level F1 of 0.8286 for Subtask 1, along with a micro-F1 of 0.7055 and a macro-F1 of 0.6496 for Subtask 2. The relatively lower performance on Subtask 2 indicates that patient-specific phenotype association remains more challenging than full-text phenotype concept recognition.

Table 4 further presents the results of the top ten teams on Leaderboard B. The first-ranked team, Agriv, achieved the highest overall score of 0.7389, followed by 医诺未来 and 天色天歌天籁音, with scores of 0.7247 and 0.7169, respectively. The performance gap among the second to tenth teams was relatively small, with overall scores ranging from 0.6987 to 0.7247, indicating a high degree of competitiveness among participating systems.

Table 4 Performance results of the top ten teams on Leaderboard B

| Rank | Men-F1 | Doc-F1 | Micro-F1 | Macro-F1 | Score |
|---|---|---|---|---|---|
| 1 | 0.7719 | 0.8286 | 0.7055 | 0.6496 | 0.7389 |
| 2 | 0.7483 | 0.8231 | 0.6890 | 0.6382 | 0.7247 |
| 3 | 0.7420 | 0.8134 | 0.6834 | 0.6289 | 0.7169 |
| 4 | 0.7521 | 0.8146 | 0.6796 | 0.6148 | 0.7153 |
| 5 | 0.7419 | 0.8051 | 0.6834 | 0.6214 | 0.7130 |
| 6 | 0.7549 | 0.8019 | 0.6813 | 0.5990 | 0.7093 |
| 7 | 0.7290 | 0.8046 | 0.6858 | 0.5981 | 0.7044 |
| 8 | 0.7482 | 0.7979 | 0.6804 | 0.5855 | 0.7030 |
| 9 | 0.7205 | 0.7807 | 0.6872 | 0.6157 | 0.7010 |
| 10 | 0.7385 | 0.7993 | 0.6572 | 0.5997 | 0.6987 |
| Baseline | 0.7211 | 0.7933 | 0.5874 | 0.5188 | 0.6552 |

Overall, 43 teams outperformed the baseline on Leaderboard B. These results demonstrate the effectiveness of the approaches developed by participating teams in improving phenotype recognition and patient-specific phenotype association. Nevertheless, the performance differences between the two subtasks highlight the remaining challenges in accurately identifying phenotype concepts associated with individual patients in full-text biomedical articles.

### 4.2 Description of Top Three Teams

The top three teams were selected based on their evaluation performance, compliance with the competition rules, and eligibility to attend the conference. Their technical approaches are summarized below.

**Team Agriv (National Science Library, Chinese Academy of Sciences)**

The first team proposed a multi-stage framework integrating phenotype candidate generation, HPO normalization, and patient-specific phenotype association. For Subtask 1, they combined three variants of PhenoTagger, a fine-tuned BioLinkBERT-large model, and HPO dictionary matching to generate candidate phenotype mentions. SapBERT was then employed to retrieve candidate HPO concepts based on semantic similarity. A two-stage LightGBM ranking framework was developed to select phenotype-HPO pairs, incorporating features from multiple recognition models, semantic similarity scores, and contextual information. A BioLinkBERT-based cross-encoder was further introduced to verify the consistency between phenotype mentions and their corresponding HPO concepts. For Subtask 2, the team constructed candidate patient-phenotype pairs and employed Qwen3-8B and DeepSeek-R1-0528-Qwen3-8B to extract patient-specific phenotype information from full-text articles. Predictions from multiple large language models, together with contextual and positional features, were integrated into a LightGBM classifier to determine patient-phenotype associations. The team also adopted five-fold cross-validation and out-of-fold predictions to improve the reliability of the training process.

**Team 医诺未来 (Yinuo Future (Sichuan) Technology Co., Ltd., Beijing Branch & Institute of Ethnology and Anthropology, Chinese Academy of Social Sciences)**

The second team developed a two-stage pipeline combining multiple phenotype recognition methods with LLM-assisted decision-making. For Subtask 1, they integrated three PhenoTagger variants based on Bioformer, PubMedBERT, and BioBERT, together with an AutoPCR-based rule method, to identify phenotype mentions and normalize them to HPO concepts. A voting strategy was adopted to retain candidate mentions identified by multiple methods, while Qwen3.5-9B was employed to resolve overlapping entities and perform negation detection and additional phenotype verification. Dictionary-based normalization and supplementary recognition of single-source candidates were also introduced to improve recognition performance. For Subtask 2, the team fine-tuned Qwen3.5-9B using two complementary strategies: direct generation of patient-associated HPO identifiers and binary classification of patient-phenotype candidate pairs. An additional LLM-based reasoning approach was designed to identify associations that might be missed by the fine-tuned models. The final patient-phenotype associations were obtained by combining predictions from these three approaches.

**Team 天色天歌天籁音 (Nanjing Tech University & Yangzhou Polytechnic University )**

The third team proposed a pipeline combining supervised phenotype recognition, semantic concept retrieval, and patient-phenotype association prediction, with an additional LLM-based verification strategy. For Subtask 1, they used the PubMedBERT branch of PhenoTagger with NegBio annotations, and added phenotype span candidates from a DeBERTa-v3-large model. SapBERT-based retrieval and CatBoost-based filtering with HPO graph features were employed for HPO normalization and candidate selection. For Subtask 2, the team combined three patient-phenotype relation scoring models, including a PubMedBERT span-relation model. Qwen3-8B was used to verify the assertion status of existing phenotype mentions. Based on the verification results, the team corrected or removed phenotype mentions and used saved relation scores to clean up or restore patient-phenotype associations. Association restoration was limited to previously scored patient candidates and used a score threshold of 0.40. This iterative refinement strategy improved the reported overall score from 0.6981 to 0.7169.

**4.3 Analysis of Top Team Solutions**

The three selected teams adopted different technical strategies for full-text phenotype recognition and patient-specific phenotype association. Table 5. summarizes their main approaches and performance on the Phase B test set.

Table 5. Comparison of the top three selected teams on Leaderboard B

| Team | Subtask 1 | Subtask 2 | Score |
|---|---|---|---|
| 1 | Multi-source candidate generation + two-stage LightGBM ranking | LLM ensemble + LightGBM | 0.7389 |
| 2 | Multi-model voting + LLM verification | Generative and discriminative fine-tuning + LLM | 0.7247 |
| 3 | Span recognition + semantic retrieval + negation correction | Association model ensemble + post-processing | 0.7169 |

As shown in Table 5, all three teams adopted multi-stage frameworks combining phenotype recognition and normalization with patient-specific association prediction. For Subtask 1, the teams integrated multiple recognition models or retrieval methods to improve phenotype candidate coverage and HPO normalization accuracy. However, their candidate selection strategies differed. The first team employed a two-stage LightGBM ranking framework with contextual cross-encoder verification, whereas the second team combined multi-model voting with LLM-assisted entity verification. In contrast, the third team emphasized supervised span recognition, semantic retrieval, and negation correction.

For Subtask 2, all three teams incorporated large language models or model ensembles to improve patient-phenotype association prediction. The first team used LLM-generated predictions as features for a LightGBM classifier, integrating semantic reasoning with contextual and positional information. The second team combined generative and discriminative fine-tuning strategies with additional LLM-assisted reasoning to improve association recall. The third team focused on refining phenotype predictions through negation verification and recovering corresponding patient-phenotype associations. These approaches demonstrate different ways of combining pretrained biomedical models, large language models, and task-specific decision mechanisms.

Among the three teams, the first team achieved the highest overall score of 0.7389, followed by The second team with 0.7247 and the third team with 0.7169. The results suggest that integrating complementary recognition models and incorporating contextual information into patient attribution can improve overall performance. In addition, the iterative correction strategy adopted by the third team highlights the importance of negation handling and consistency between phenotype recognition and patient-specific association prediction. Overall, these findings indicate that effective integration of biomedical language models, LLM-based reasoning, and task-specific post-processing remains an important direction for patient-specific phenotype information extraction from full-text biomedical articles.

## 5 Conclusion

In this paper, we presented an overview of CHIP 2026 Shared Task 1, focusing on patient-specific phenotype concept recognition from full-text biomedical literature. The task established a standardized evaluation framework for full-text phenotype recognition, HPO normalization, and patient-specific phenotype association using 200 annotated full-text biomedical articles from the PatientPheX dataset. The high level of participation and diverse technical approaches demonstrated growing research interest in automated phenotype information extraction. The top-performing teams adopted multi-stage frameworks combining biomedical pretrained language models, large language models, and ensemble strategies to improve phenotype recognition and patient-phenotype association prediction. The best-performing system achieved an overall score of 0.7389 on the final test set, demonstrating the effectiveness of integrating complementary models and contextual information for patient-specific phenotype extraction.

Despite these advances, several challenges remain. Accurately recognizing diverse phenotype expressions and mapping them to standardized HPO concepts remains difficult, particularly for negated and composite phenotype. Furthermore, patient-specific phenotype association requires distinguishing clinical findings among multiple patients and family members, often involving long-range contextual information across different sections of full-text articles. These challenges highlight the need for more effective methods that jointly model phenotype recognition and patient attribution. For future work, we plan to explore more effective methods for patient-specific phenotype information extraction, particularly in addressing complex phenotype expressions, patient attribution, and long-range contextual dependencies in full-text biomedical literature. These efforts aim to improve the accuracy and robustness of patient-specific phenotype extraction and support downstream applications, such as patient-specific knowledge base construction, genotype-phenotype association analysis, and rare disease diagnosis.

## Acknowledgements

This work was supported by the National Natural Science Foundation of China (No. 62302076, 62276043, 62676065) and the Fundamental Research Funds for the Central Universities (No. DUT25YG108).